\documentclass[conference]{IEEEtran}
\IEEEoverridecommandlockouts

\usepackage{cite}
\usepackage{amsmath,amssymb,amsfonts}
\usepackage{algorithmic}
\usepackage{graphicx}
\usepackage{textcomp}
\usepackage{xcolor}
\usepackage{flushend}
\usepackage{booktabs}
\usepackage{array} 

\def\BibTeX{{\rm B\kern-.05em{\sc i\kern-.025em b}\kern-.08em
    T\kern-.1667em\lower.7ex\hbox{E}\kern-.125emX}}
\begin{document}

\title{Hallucination Detection with Small Language Models}

\author{
\IEEEauthorblockN{Ming Cheung}
\IEEEauthorblockA{\textit{dBeta Labs, The Lane Crawford Joyce Group} \\
Hong Kong, China \\
mingcheung@lcjgroup.com}
}

\maketitle

\begin{abstract}
Since the introduction of ChatGPT, large language models (LLMs) have demonstrated significant utility in various tasks, such as answering questions through retrieval-augmented generation. 
Context can be retrieved using a vectorized database, serving as a foundation for LLMs to generate responses. 
However, hallucinations in responses can undermine the reliability of LLMs in practical applications, and they are not easily detectable in the absence of ground truth, particularly in question-and-answer scenarios.
This paper proposes a framework that integrates multiple small language models to verify responses generated by LLMs using the retrieved context from a vectorized database. 
By breaking down the responses into individual sentences and utilizing the probability of generating "Yes" tokens from the outputs of multiple models for a given set of questions, responses, and relevant context, hallucinations can be detected.
The proposed framework is validated through experiments with real datasets comprising over 100 sets of questions, answers, and contexts, including responses with fully and partially correct sentences. 
The results demonstrate a 10\% improvement in $F1$ scores for detecting correct responses compared to hallucinations, indicating that multiple small language models can be effectively employed for answer verification, providing a scalable and efficient solution for both academic and practical applications.
\end{abstract}


\section{Introduction}
The development of large language models (LLMs) has revolutionized various fields, offering remarkable capabilities in natural language understanding and generation. 
However, despite their impressive performance, LLMs are prone to generating inaccurate information, commonly referred to as "hallucinations." 
Table \ref{tab:Contradiction} presents examples of common types of hallucinations found in responses to prompts: Logical, Prompt, and Factual contradictions. 
The Logical contradiction illustrates a scenario where the city of Madison is misrepresented regarding its population, claiming that a town with a population of 500K is not a small town. 
The Prompt contradiction showcases a misleading description of a healthy breakfast that contradicts common dietary guidelines. 
Lastly, the Factual contradiction highlights an inaccurate ingredient in a traditional Margherita pizza, which incorrectly lists chocolate as a key component. 
These inaccuracies can undermine the reliability of LLMs, particularly in high-stakes applications where precision is essential.

Hallucinations in LLMs arise from various factors, including source-reference divergence in the training data, which may result from heuristic data collection methods or the inherent complexities of certain natural language generation tasks \cite{huang2023survey}.

\begin{table}[t]
    \centering
    \caption{Contradiction Types and their Examples}
    \label{tab:Contradiction}
    \begin{tabular}{@{}p{2cm} p{2cm} p{3.5cm}@{}}
        \toprule
        \textbf{Hallucination Types} & \textbf{Prompt} & \textbf{Generated Response} \\ 
        \midrule
        Logical & Can you introduce Madison? & The city of Madison has over 500K residents. It is known for its small-town charm and quiet atmosphere. \\ 
        \midrule
        Prompt & Describe a healthy breakfast that includes fruits and whole grains. & A bowl of sugary cereal with milk and a side of bacon is a great choice for breakfast. \\ 
        \midrule
        Factual & What are the main ingredients in a traditional Margherita pizza? & A traditional Margherita pizza is made with tomatoes, mozzarella cheese, fresh basil, and a secret ingredient: chocolate. \\ 
        \bottomrule
    \end{tabular}
\end{table}
\begin{figure*}
\centering 
\includegraphics[width=7in]{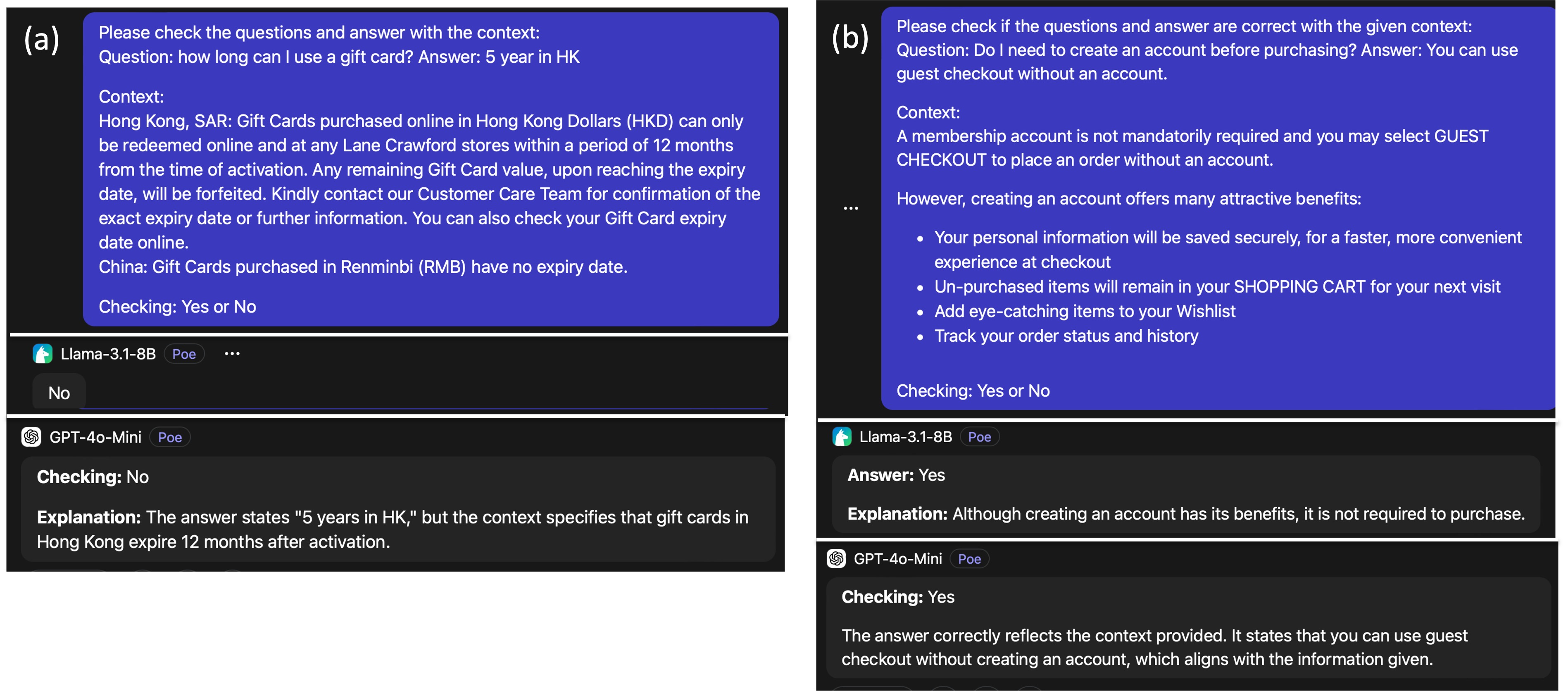} 
\caption{Examples of multiple choices for checking responses.}
\label{fig:prompt_MC}
\end{figure*}
Detecting hallucinations in LLM-generated responses is both critical and challenging. 
Traditional LLM performance measurements, such as ROUGE \cite{lin2004rouge}, are not suitable for real-time evaluation, as the necessary reference answers are not readily available for detection.
An alternative approach leverages the fact that language models are trained to answer yes or no questions \cite{kadavath2022language}. 
This method prompts an LLM to answer "yes" or "no" in response to a given set of questions and answers to detect hallucinations in generated responses.
It utilizes the knowledge of the LLM, though this knowledge is not always accessible.
The capabilities of these language models can be enhanced with retrieval-augmented generation (RAG) \cite{lewis2020retrieval} using vectoried database \cite{jing2024large}.
Even if the information is not in the language model, it can utilize RAG context to verify responses from other LLMs.
One can call an LLM multiple times, similar to an API, to obtain probability estimates, but this requires more time.
Deploying the model locally enables the computation of "yes" and "no" probabilities without multiple calls, allowing for direct extraction of probabilities.
However, due to the models being closed-source, such as ChatGPT, or their large size requiring significant computational resources to host, this is not always feasible.
In contrast, smaller language models (SLMs), defined as having 100M to 5B parameters \cite{lu2024small}, present a promising alternative for answer verification. 
While these models may not match the performance of their larger counterparts in generating free-text responses, they can be more efficient and cost-effective for specific tasks. 
Figure \ref{fig:prompt_MC} illustrates examples of checking with questions, answers, and contexts.
It has been observed that even a relatively small model can still produce accurate results.

This paper proposes a framework for detecting hallucinations by exploring the potential of integrating SLMs for contextual answer checking, aiming to improve verification accuracy while mitigating the resource demands associated with larger models. 
The main contributions of this paper are:
\begin{itemize}
  \item A framework for computing hallucination scores using multiple SLMs;
  \item An algorithm for the framework to compute scores from segmented responses using different SLMs;
  \item A dataset prepared to verify the proposed framework, demonstrating an improvement of 10\% over the baseline.
\end{itemize}

The paper is organized as follows: Section \ref{sec:related_works} discusses related works, and Section \ref{sec:qanda} introduces how questions can be handled with retrieval-augmented generation in LLMs. 
Section \ref{sec:proposed_flow} discusses the proposed framework, while Section \ref{sec:experiments} covers the experimental settings and results. 
Finally, Section \ref{sec:conclusion} concludes the paper.

\section{Related Works}
\label{sec:related_works}
\noindent
The emergence of large language models (LLMs) has significantly transformed multiple domains, providing extraordinary capabilities in natural language understanding and generation. LLMs demonstrate substantial proficiency across a variety of NLP tasks, including language translation \cite{reynolds2021prompt}, sentiment analysis, named entity recognition \cite{kaddour2023challenges, xie2023translating}, and SQL generation, among others \cite{cheung2024reality}. Historical models, such as recurrent neural networks (RNNs) \cite{salehinejad2017recent} and long short-term memory (LSTM) models \cite{graves2012long}, laid the foundational understanding of sequential data, particularly in text processing. However, the advancement of the transformer architecture \cite{gillioz2020overview} marked a pivotal turning point, enabling models to effectively capture long-range dependencies within textual data. This architecture employs self-attention mechanisms \cite{vaswani2017attention}, allowing the model to concentrate on disparate segments of the input text, thereby generating coherent and contextually appropriate responses. Consequently, numerous LLMs have been developed, including GPT-3 \cite{dale2021gpt}, Google Bard \cite{manyika2023overview}, and NeevaAI \cite{liu2023evaluating}. These models, refined through human feedback \cite{ouyang2022training}, encompass billions of parameters and are capable of engaging in conversational interactions \cite{brown2020language}.

\indent
LLMs have drawn attention for their capabilities; however, they are known to produce hallucinations in their outputs, and there is no clear definition of hallucination \cite{venkit2024confidently}. Detecting hallucinations is not analogous to conventional LLM measurements, such as the ROUGE metric \cite{guo2017calibration} and BLEU score \cite{papineni2002bleu}, which measure response quality by comparing the output with a ground truth. The difficulty in detecting hallucinations arises from the absence of ground truth during response generation. Another approach involves building a classifier to evaluate the response \cite{chen2023hallucination, filippova2020controlled}. The hidden states in the models are also useful for constructing classifiers \cite{orgad2024llms}. Reinforcement learning can be applied to detect hallucinations in logic \cite{havrilla2024glore} and to correct them \cite{welleck2022generating}.

\indent
Another solution is to prompt LLMs to answer yes or no questions \cite{kadavath2022language} to detect hallucinations in a response. The research in \cite{farquhar2024detecting} employs entropy and sentence splitting to generate multiple outputs. Additionally, the knowledge required for verification may not exist within the LLM, necessitating retrieval-augmented generation (RAG) \cite{lewis2020retrieval} to enhance its capacity by providing knowledge that does not exist in the LLM. 
Related context can be extracted from vectorised database \cite{jing2024large}.
Based on the context, score-based hallucination tests utilize distribution tests approximated solely by the answer \cite{yadkori2024believe}. However, this approach necessitates multiple rounds of generation and precise probability computations, making it challenging to apply to large LLMs that are either not open (e.g., ChatGPT) or excessively large. The work in \cite{kadavath2022language} investigates the probability of correctness, \( P(\text{True}) \), by estimating the likelihood of a "yes" response, leveraging the strengths of LLMs in predicting answers for multiple-choice questions \cite{sprague2024cot}. 

This paper proposes a framework to utilize multiple small language models (SLMs) for detecting hallucinations with \( P(\text{True}) \).
\begin{figure*}
\centering 
\includegraphics[width=7in]{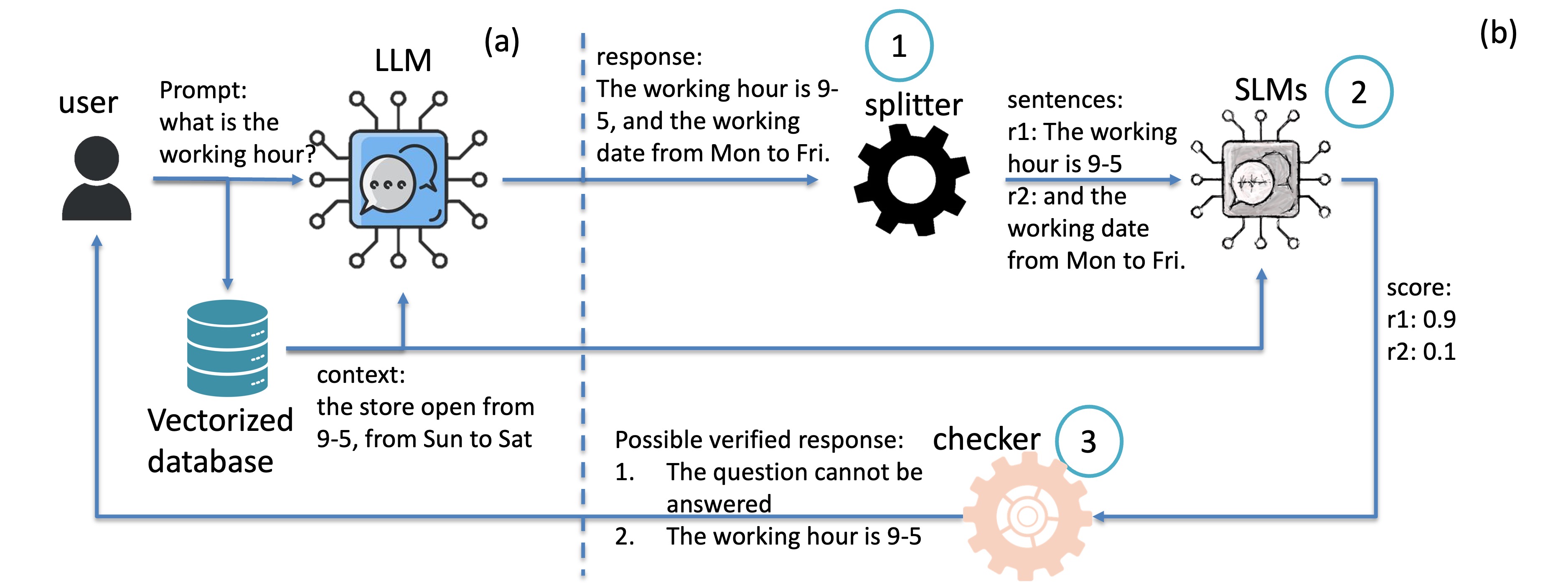} 
\caption{Flow of using the proposed framework: (a) LLMs generate responses; (b) the proposed framework.}
\label{fig:proposedFlow}
\end{figure*}

\section{Questions and Answering using RAG}
\label{sec:qanda}
This section discusses how responses can be generated using RAG, as shown in Fig. \ref{fig:proposedFlow} (a). The first part discusses some common Large Language Models, followed by RAG.

\subsection{Large Language Model (LLM)}
\noindent
The Large Language Model (LLM) serves as the primary engine for processing user queries. The LLM leverages vast datasets to understand context and provide accurate, human-like answers, making it a crucial component of the interaction. The prompt may include the role, the questions, as well as the required context to answer the questions. Even with a prompt containing suitable context, there is no guarantee that the responses will be free of hallucinations.

\indent
Some common LLMs include ChatGPT 3.5\footnote{https://openai.com/}, which is a conversational AI model developed by OpenAI, utilizing the GPT (Generative Pre-trained Transformer) architecture\cite{radford2018improving}. It generates human-like responses for various tasks, trained on 570 gigabytes of text from sources like Wikipedia and Twitter, with 175 billion parameters. Another popular LLM is Meta’s Llama-2-70b\footnote{https://huggingface.co/meta-llama/Llama-2-70b-chat-hf}, which employs the GPT-3.5 architecture and is designed for high-quality response generation across applications like content creation and natural language understanding. It features 70 billion parameters and is trained on 2 trillion tokens from publicly available sources\cite{huggingFace}. APIs are available for different applications, allowing users easier access to these models for generating responses to their questions.

\subsection{Retrieval Augmented Generation}
\noindent
There are many different LLMs available, and due to variations in model structures, training data, and decoding strategies, they can generate differ ent responses for the same prompt and question. Additionally, users may want to answer questions based on specific documents, such as inquiries from company employees where the documents are staff handbooks. Although one can fine-tune a model to respond accordingly\cite{hu2021lora}, this requires substantial resources and expertise to train and host the model. 

One solution is to provide related context for a question, known as Retrieval Augmented Generation \cite{bechard2024reducing}. As the models are trained to follow human instructions \cite{ouyang2022training}, LLMs can generate responses based solely on the provided context. The goal of the next section is to discuss how the response can be verified for hallucinations.

\section{Proposed Flow}
\label{sec:proposed_flow}
\noindent
The flow diagram, as shown in Fig. \ref{fig:proposedFlow}, outlines the proposed framework, from the LLM generating responses to returning a hallucination score for that response. This response is first processed by a splitter, which divides the information into two separate streams: one confirming the accuracy of the details and the other assessing the context. The system subsequently checks if all conditions are met, determining whether the working hours align with the specified criteria. Hence, the hallucination score, $s^m_i$, can be computed as:
\begin{equation}
\label{eq:f1}
s^{(m)}_i = P(h_i=1| q_i, r_i, c_i, prompt)
\end{equation}
where $s^m_i$ is the score for response $i$, while $m$ is the $m$-th SLM. The value $h_i$ indicates hallucination, and $prompt$ is the prompt for the check. The prompt asks the SLM to generate a "yes" or "no" answer, as shown in Fig. \ref{fig:prompt_MC}. The parameters $q_i$, $r_i$, and $c_i$ represent the question, response, and context, respectively.

By considering the first generated token, Eq. \ref{eq:f1} becomes:
\begin{equation}
\label{eq:f2}
s^{(m)}_i = P(token_{1}=\mathbf{yes}| q_i, r_i, c_i)
\end{equation}
where $token_{1}$ is the first token generated by the SLM, and $s^m_i$ is the probability that the first token is $\mathbf{yes}$. Note that $prompt$ is omitted as all SLMs use the same prompt. This section introduces each part of the flow one by one.

\subsection{Splitter}
\noindent
The Splitter acts as an intermediary that takes the LLM's response and divides it into sentences for further processing. It separates critical details, such as the working hours and contextual information shown in Fig. \ref{fig:proposedFlow}, to ensure that each piece of information can be individually evaluated. For a response, $r_i$, it will be split into several sub-responses, denoted as $r_{i, j}$. Without this step, evaluating the whole sentence with both correct and incorrect information would confuse the checker. One method to split a response is using SpaCy\footnote{spacy.io}, a powerful NLP library in Python, which effectively segments text into sentences by identifying the relationships between words and punctuation marks. The response, $r_i$, can be computed by splitting $r_i$ into $r_{i,j}$.

\subsection{Small Language Models (SLMs)}
\noindent
The Small Language Model (SLM) is responsible for evaluating the segmented response. It checks for accuracy, such as the working hours and operational days, against the context. The question, context, and answer are given in the prompt, and the SLM is asked to generate a response starting with YES or NO\cite{kadavath2022language}. The score for a split sentence, $s^{(m)}_{i,j}$, is computed as:
\begin{equation}
\label{eq:f3}
s^{(m)}_{i,j} = P(token_{1}=\text{yes}| q_i, c_i, r_{i,j}) 
\quad \textbf{where} \quad r_{i,j} \subset S(r_i) 
\end{equation}
The checking is conducted on each split sentence. There may be more than one SLM in the validation, and the score from each SLM is sent to the checker. SLMs, such as Qwen2 \cite{bai2023qwen} and MiniCPM \cite{hu2024minicpm}, demonstrate good performance \cite{lu2024small} across many tasks, including math and problem-solving, making them suitable candidates for the framework.

\subsection{Checker}
\noindent
The checker is the final decision-making component that combines the scores from the split sentences, processed by multiple SLMs, into a single score. By confirming or rejecting the details based on the scores, the answers can be evaluated for accuracy, enhancing the reliability of the overall system. Different SLMs have different scales, meaning they possess varying means and variances for the same set of data. Consequently, the values of the responses from different SLMs are normalized as:
\begin{equation}
\label{eq:normalization}
\tilde{s}^{(m)}_{i,j} = \frac{s^{(m)}_{i,j} - \mu_m}{\sigma_m}
\end{equation}
where $\mu_j$ and $\sigma_j$ are the means and variances of the SLM $j$. These can be computed based on previous responses. 

The score, $s_{i,j}$, for a segmented response $j$, based on multiple SLMs, can be calculated as:
\begin{equation}
\label{eq:f4}
s_{i,j} = \frac{1}{M}\sum^{M} \tilde{s}^{(m)}_{i,j}
\end{equation}
where $M$ is the total number of SLMs. The final score, $s_i$, of response $i$, can be computed using the harmonic mean:
\begin{equation}
s_i = \frac{|S(r_i)|}{\sum_{j=1}^{|S(r_i)|} \frac{1}{s_{i,j}}}, \quad s_{i,j} > 0 
\label{eq:harmonic}
\end{equation}
To avoid issues with non-positive values, any values less than or equal to zero are adjusted. More discussions regarding the choice of means can be found in later sections.

\section{Experiments and Measurement}
\label{sec:experiments}
This study conducts experiments and measurements on a real dataset sourced from Lane Crawford. The dataset comprises questions related to Human Resources policies, with corresponding contexts extracted from the employee handbook.
\begin{figure*}
\centering 
\includegraphics[width=7in]{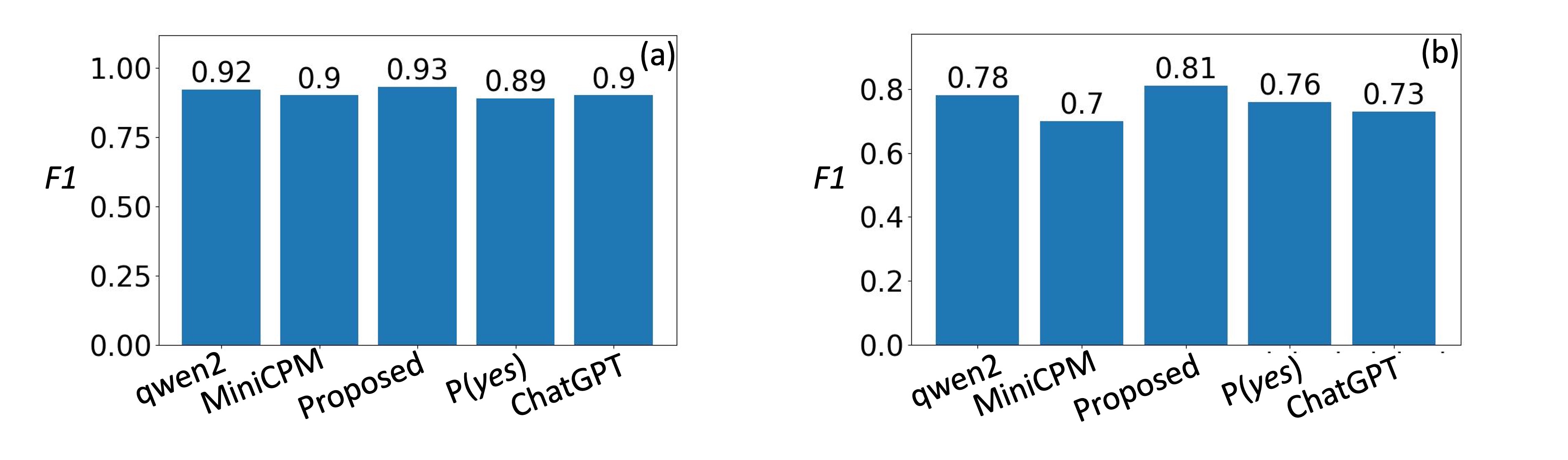}
\caption{Experiment results on the best $F1$ for different approaches in detecting correct responses from: (a) wrong, (b) partial.}
\label{fig:proposed}
\end{figure*}
\begin{figure*}
\centering 
\includegraphics[width=7in]{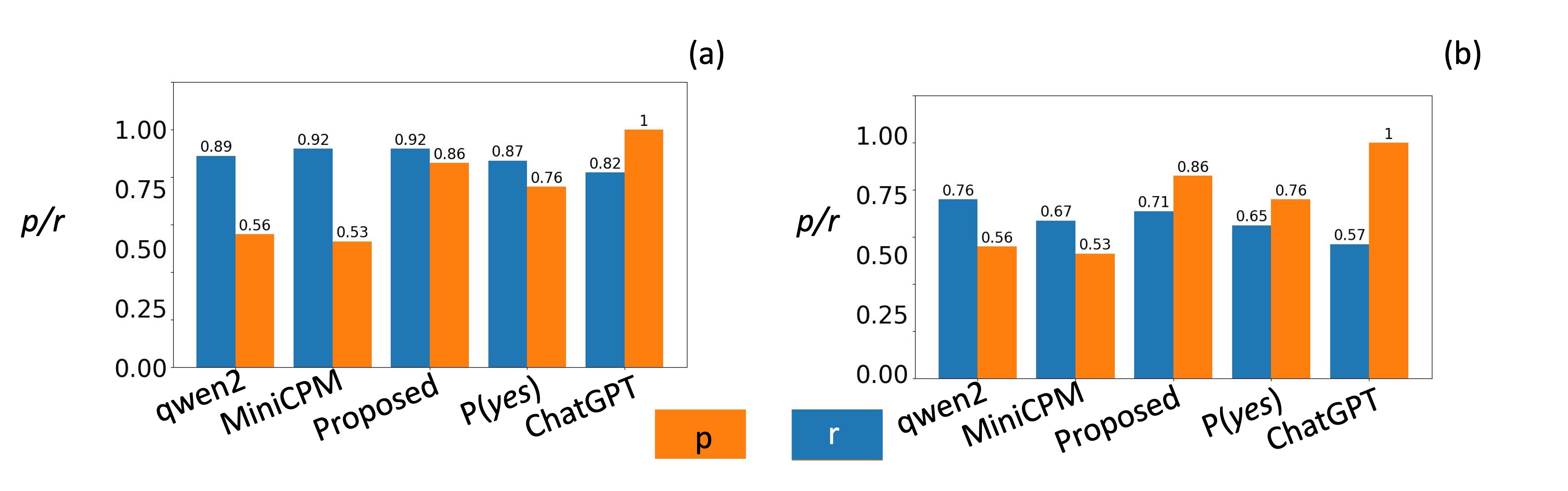}
\caption{Experiment results on the best $p$ and $r$ for detecting correct responses from: (a) wrong, (b) partial.}
\label{fig:proposed_p_r}
\end{figure*}
\subsection{Dataset}
\noindent
The dataset is generated from the handbook of Lane Crawford. The questions are formulated based on the context of various topics from the handbook, ranging from Employment (such as probation, salary, leave, and benefits) to Policy (such as uniform and emails), as well as other matters (such as handling media requests and bringing personal devices to work). The following is an example of context and the question regarding working hours, derived from the context:
\begin{itemize}
    \item Context: The store operates from 9 AM to 5 PM, from Sunday to Saturday. There should be at least three shopkeepers to run a shop.
    \item Question: What are the working hours?
\end{itemize}
Based on the context and questions, three responses are generated for each pair of context and question. Note that the context may contain more information than is necessary to formulate the question. The responses belong to three different categories: one labeled as correct, one as partially correct, and one as incorrect. The labeled responses are as follows:
\begin{itemize}
    \item Correct: The working hours are 9 AM to 5 PM, and the store is open from Sunday to Saturday.
    \item Partial: The working hours are 9 AM to 5 PM, and the store is open from \textit{Monday to Friday}.
    \item Wrong: The working hours are 9 AM to \textit{9 PM}, and you \textit{do not need to work on weekends}.
\end{itemize}
The incorrect parts are italicized. The reason the response is labeled as partial is that while the working hours are accurate, the days are not. Each question in the dataset is accompanied by a context that contains the essential information needed to formulate an answer and includes three responses labeled as “correct,” “partial,” and “incorrect.” It is important to note that the labels are not applied at the sentence level, particularly in the case of partial responses. The primary aim of these experiments is to assess whether the SLM can effectively identify incorrect or partial responses in comparison to the correct ones. Using the same contexts and questions to form multiple answers ensures that the models are not biased toward certain contexts.

\subsection{Small Language Models}
Two models are used as the SLM in the proposed framework: Qwen2 \cite{bai2023qwen} and MiniCPM \cite{hu2024minicpm}. This section introduces them one by one.

\subsubsection{Qwen2}
Qwen2 \cite{bai2023qwen} is an advanced Language Model (LLM) designed to enhance natural language understanding and generation\footnote{https://huggingface.co/Qwen/Qwen2.5-1.5B-Instruct}. Building on the capabilities of its predecessors, Qwen2 features improved contextual awareness, enabling it to generate more coherent and contextually relevant responses. It utilizes a vast dataset for training, allowing it to perform a variety of tasks, from conversational agents to content creation\footnote{available: huggingface.co/Qwen/Qwen2-1.5B-Instruct}.

\subsubsection{MiniCPM}
MiniCPM \cite{hu2024minicpm} is a series of edge-optimized large language models designed for efficient deployment on devices with limited computational resources\footnote{https://huggingface.co/openbmb/MiniCPM-2B-sft-bf16}. The flagship model, MiniCPM-2B, features 2.4 billion parameters and excels in tasks such as language understanding and coding, often outperforming larger models like Llama2-13B and MPT-30B. MiniCPM also supports extensive customization, allowing users to fine-tune outputs for various NLP tasks, thus broadening its applicability in real-time and on-device scenarios.

\begin{figure*}
\centering 
\includegraphics[width=7in]{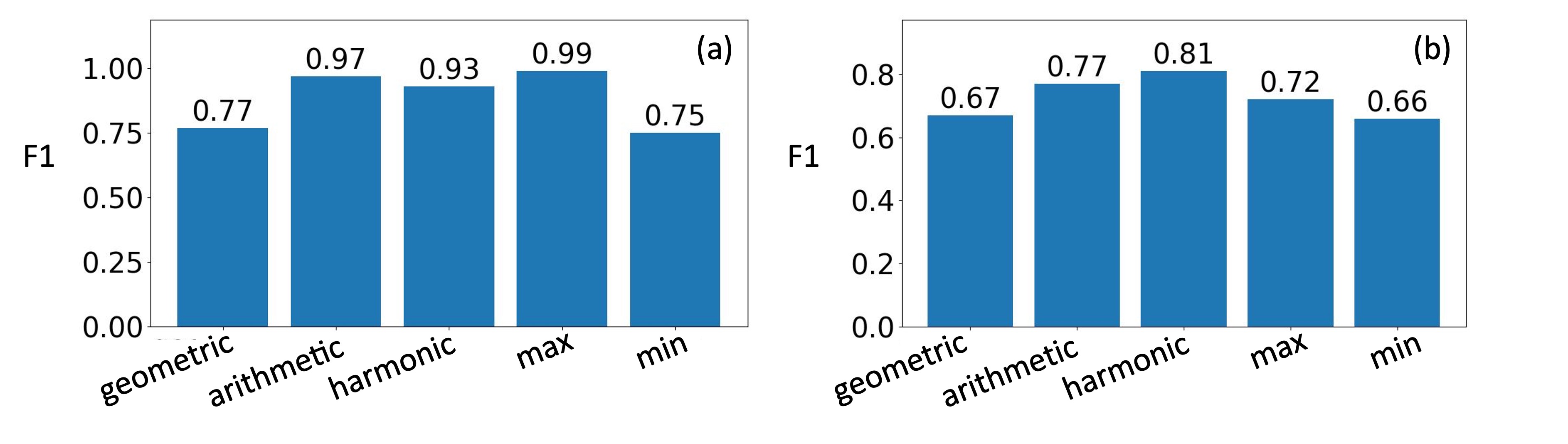}
\caption{Experiment results on different means algorithms in detecting correct responses from: (a) wrong, (b) partial.}
\label{fig:MeansProposed}
\end{figure*}
\subsection{Approaches and Baselines}
This section introduces the baselines and the approaches for the frameworks. 
\begin{itemize}
    \item Proposed: Using Qwen2 and MiniCPM as the SLMs in the proposed framework;
    \item ChatGPT: Prompt ChatGPT to answer "Yes" or "No" ($P(\text{True})$) \cite{kadavath2022language};
    \item $P(\text{yes})$: Prompt SLM for the whole response for the probability to be "YES" using Qwen2;
    \item Qwen2: Similar to the proposed, but using only Qwen2 as the SLM.
    \item MiniCPM: Similar to the proposed, but using only MiniCPM as the SLM.
\end{itemize}
The approach using ChatGPT is for those LLMs that are only available through API, while $P(\text{yes})$ is the approach without a splitter. Approaches using Qwen2 and MiniCPM demonstrate the effectiveness of utilizing multiple SLMs.

\subsection{Results}
An experiment is conducted to compare the $F1$ scores of different approaches in detecting "correct" responses from "partial" or "wrong" ones. If the score in Eq. \ref{eq:harmonic} exceeds a threshold, the response is labeled as "correct"; otherwise, it is not. The results are shown in Fig. \ref{fig:proposed}. Fig. \ref{fig:proposed} (a) shows the results of detecting "correct" responses from "wrong" responses, while Fig. \ref{fig:proposed} (b) illustrates detecting "correct" responses from "partial" responses, with the thresholds yielding the highest $F1$ scores selected. In Fig. \ref{fig:proposed} (a), all approaches achieve high $F1$ scores, with $P(\text{yes})$ being the lowest at 0.89. The results indicate that all approaches can detect responses that are completely contradictory to the contexts. Fig. \ref{fig:proposed} (b) shows lower $F1$ scores, with the proposed method achieving the highest score of 0.81. These results suggest a general decrease in performance from detecting "wrong" to detecting "partial" responses, as the latter is much more challenging. Utilizing SLMs is better than using ChatGPT and $P(\text{yes})$ by 11\% and 6.6\%, respectively. Additionally, employing multiple SLMs outperforms the use of a single model.

A second test is conducted to measure the best precision $p$ and the corresponding recall $r$ of different approaches. As a question-and-answering system, it is desirable to have a high $p$ and reasonable $r$, meaning a system that answers only those questions it is confident about, without providing incorrect information. Fig. \ref{fig:proposed_p_r} shows the $p$ and $r$ for different approaches, detecting correct responses from wrong versus detecting correct responses from partial, in Fig. \ref{fig:proposed_p_r} (a) and Fig. \ref{fig:proposed_p_r} (b), respectively. Note that $r$ must be at least 0.5 while selecting the $p$, to prevent selecting a very high $p$ with a very low $r$. In Fig. \ref{fig:proposed_p_r} (a), the $p$ score for "Qwen2" is 0.89, with MiniCPM at 0.92. However, they have relatively low $r$ values of 0.56 and 0.53, respectively. Conversely, the proposed method has a comparable $p$ but a much higher $r$, implying that using multiple SLMs could enhance performance. Similarly, Fig. \ref{fig:proposed_p_r} (b) shows the results for detecting correct responses from partial responses. All approaches exhibit lower $p$ and $r$, yet the same conclusion can be drawn: utilizing multiple SLMs improves performance, and the proposed method is superior to both $P(\text{yes})$ and ChatGPT.

\begin{figure*}
\centering 
\includegraphics[width=7in]{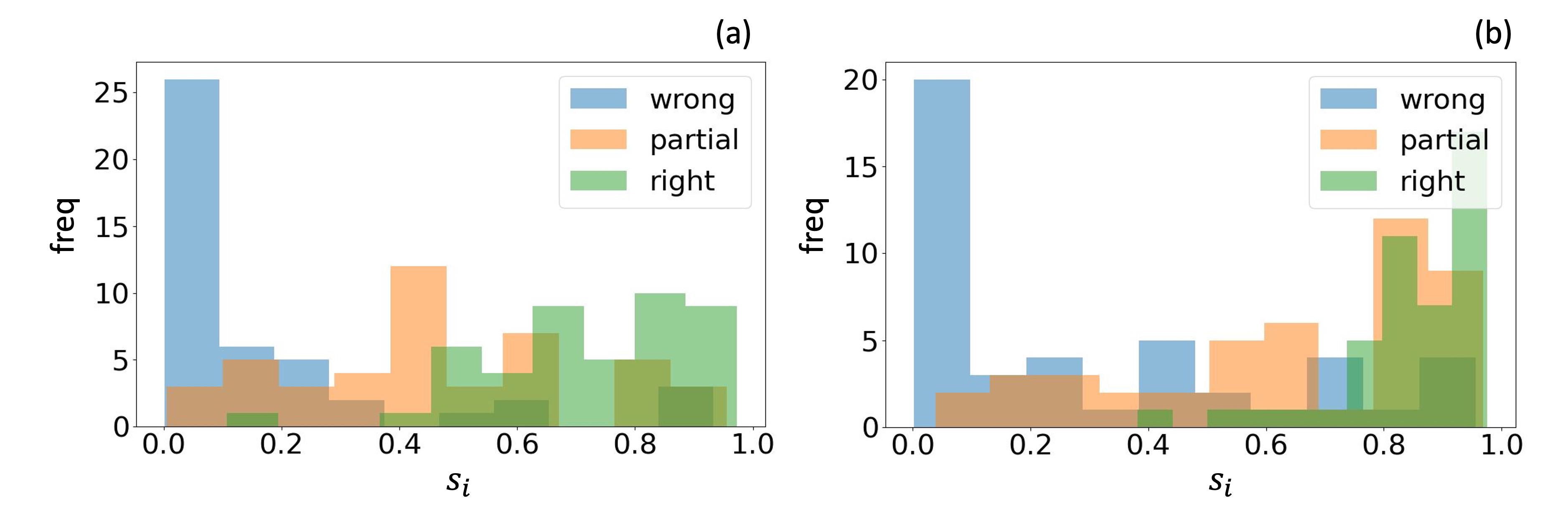}
\caption{Distributions of approaches: (a) proposed, (b) $P(\text{yes})$.}
\label{fig:distributions}
\end{figure*}

\begin{figure*}
\centering 
\includegraphics[width=7in]{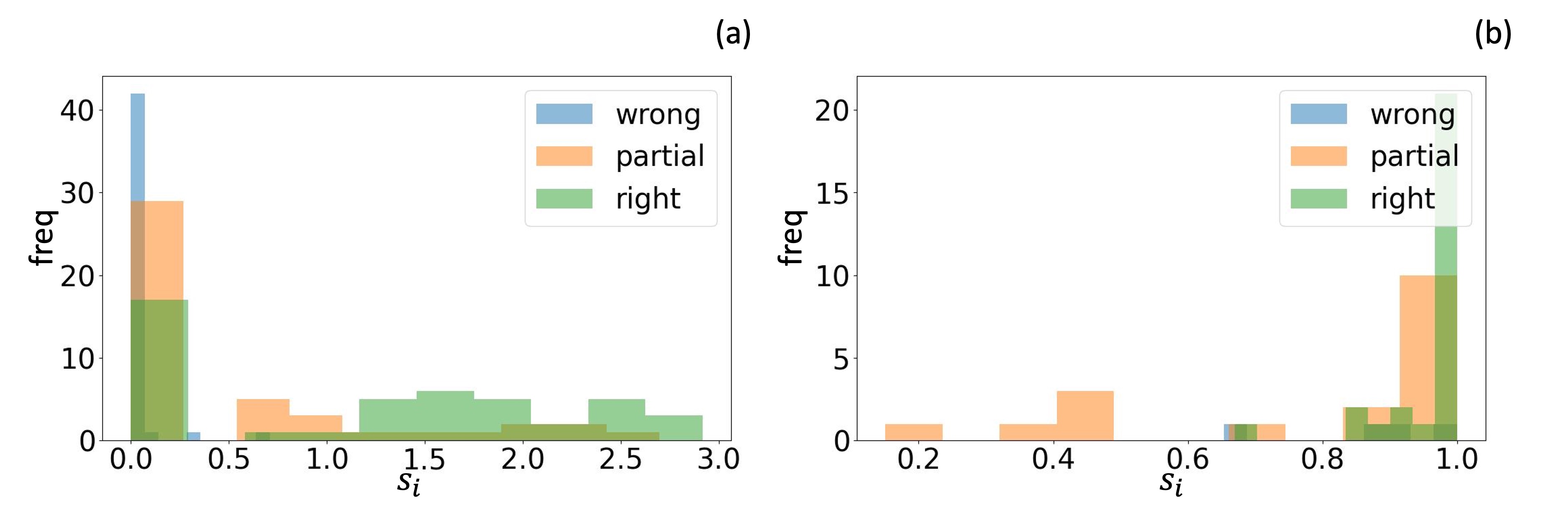}
\caption{Distributions of two means: (a) geometric, (b) harmonic.}
\label{fig:distributions_2models}
\end{figure*}

\subsection{Means}
It is interesting to investigate how the mean calculation affects the results. Fig. \ref{fig:MeansProposed} presents two bar charts comparing the $F1$ scores across different methods labeled as "geometric," "arithmetic," "max," and "min." The calculations for these means are as follows:
\begin{equation}
s_i(S, \text{m = 'ari'}) = \frac{1}{|S(r_i)|} \sum_{j=1}^{|S(r_i)|} s_{i,j} \label{eq:arithmetic}
\end{equation}
where $|S(r_i)|$ is the number of elements in $S$. The geometric mean can be calculated as:
\begin{equation}
s_i(S, \text{m = 'geo'}) = \exp\left(\frac{1}{|S(r_i)|} \sum_{j=1}^{|S(r_{i})|} \log(s_{i,j})\right), \quad s_{i,j} > 0
\label{eq:geometric}
\end{equation}
To find the minimum value in the dataset, we use:
\begin{equation}
s_i(S, \text{m = 'min'}) = \min(S(r_i)) \label{eq:min}
\end{equation}
This represents the minimum value among all sentences. The maximum value is calculated as:
\begin{equation}
s_i(S, \text{m = 'max'}) = \max(S(r_i)) \label{eq:max}
\end{equation}
This represents the maximum value among all sentences. Hence, Fig. \ref{fig:MeansProposed} shows the results of testing different mean calculations. In Fig. \ref{fig:MeansProposed} (a), the $F1$ scores range from 0.75 to 0.99, with the highest score observed for the "max" method (0.99). Fig. \ref{fig:MeansProposed} (b) shows lower $F1$ scores, with the highest score of 0.81 for the "harmonic" method and the lowest at 0.66 for the "min" method. It is observed that the "max" method does not work well for "partial" responses, while there are good correct and hallucination sentences in one response. The results indicate that the harmonic mean yields the best outcomes.

\subsection{Distributions}
It is interesting to investigate why and how the proposed method works. Hence, Fig. \ref{fig:distributions} presents two histograms illustrating the frequency of responses categorized as "wrong," "partial," and "correct" across different values of $s_i$. Fig. \ref{fig:distributions} (a) and (b) show the histograms of $s_i$ for the proposed method and $P(\text{yes})$, respectively. In both figures, "wrong" responses yield lower values of $s_i$, while "correct" responses consistently show higher $s_i$ values and a higher frequency at elevated values of $s_i$. The "partial" responses are consistently present across both histograms but are less frequent than the "wrong" and "correct" categories. The "partial" responses spread between the two categories, which is one reason that partial responses are difficult to detect using $s_i$. Fig. \ref{fig:distributions} (a) exhibits a more pronounced peak for "wrong" responses compared to Fig. \ref{fig:distributions} (b). Although most "correct" responses in $P(\text{yes})$ have a high $s_i$, they are inseparable from "partial" responses, while the proposed method can distinguish them and achieve better performance.

\indent
A similar conclusion can be drawn from using different means. Fig. \ref{fig:distributions_2models} (a) and (b) present two histograms illustrating the frequency distribution of correctness categories (wrong, partial, correct) against the variable $s$ for geometric and harmonic means. In Fig. \ref{fig:distributions_2models} (a), there is a high frequency of "wrong" responses at lower values of $s$, while "partial" and "correct" responses show lower frequencies. Additionally, most "correct" responses have high values. Fig. \ref{fig:distributions_2models} (b) indicates that the "correct" responses are found at higher values of $s$. The overall trend suggests that as $s$ increases, the frequency of correct responses rises while the frequency of wrong responses decreases, highlighting a positive correlation between $s$ and correctness. Note that Fig. \ref{fig:distributions_2models} (b) only shows responses with values greater than 0, and more "wrong" responses are not depicted.

\section{Conclusion}
\label{sec:conclusion}
\noindent
This paper presents a framework for utilizing SLMs in answer verification for a given set of context, answers, and questions. 
A real dataset is created to demonstrate their effectiveness in ensuring contextual relevance in responses. 
The experiments show that the proposed approach is 11\% and 6.6\% better than those using ChatGPT and P(yes), by utilizing two SLMs, Qwen2 and MiniCPM, within the proposed framework. 
The findings suggest that SLMs are a viable alternative for applications demanding efficient and scalable answer verification, particularly in resource-constrained environments. 
Future research could focus on optimizing the framework's performance on different types of data or on better integration of SLMs, such as adding gating mechanisms \cite{zhou2022mixture}. 
Another direction is to integrate with verification frameworks \cite{chen2023complex} to extract additional information online for checking general context.

\bibliographystyle{IEEEtran}
\bibliography{llm}
\end{document}